# ForEx: A Formal Verification Framework for Explainable Reasoning in Logical Fallacy Detection and Annotation

Pei-Cing Huang
*Department of Information Management*
*National Sun Yat-Sen University*
Kaohsiung, Taiwan
pcpeicing@gmail.com

Chienyu Liu
*Department of Information Management*
*National Sun Yat-Sen University*
Kaohsiung, Taiwan
c09010011@gmail.com

Chan Hsu
*Department of Information Management*
*National Sun Yat-Sen University*
Kaohsiung, Taiwan
chanshsu@gmail.com

Ci-Siang Chen
*Department of Information Management*
*National Sun Yat-Sen University*
Kaohsiung, Taiwan
cisiang.chen@gmail.com

Pei-Ju Lee
*Graduate Institute of Data Science & Information Computing*
*National Chung Hsing University*
Taichung, Taiwan
pjlee@nchu.edu.tw

Yihuang Kang
*Department of Information Management*
*National Sun Yat-Sen University*
Kaohsiung, Taiwan
ykang@mis.nsysu.edu.tw

***Abstract*—Current evaluations of Large Language Models (LLMs) on logical fallacy detection focus on predicted labels, but do not establish whether those labels are supported by the reasoning the models provide. We propose ForEx (Formal Verification for Explainable Reasoning), a framework that translates LLM-generated explanations into Lean4 and verifies whether the translated rationale is derivable under encoded premises, not the logical validity of the original natural language argument. To distinguish prediction outcomes from the formal status of the supporting reasoning, we introduce the LLM Argument Verification Matrix, which separates label consistency from formal verification status. Experiments on LOGIC-Climate show that over 90% of LLM outputs can be translated into formal reasoning chains that pass verification, while agreement with human annotations remains around 20%. These results expose a systematic gap between formal derivability and label agreement, a distinction invisible to prediction-based metrics. ForEx moves LLM evaluation beyond label correctness toward machine-checkable analysis of formalized reasoning chains.**

## I. Introduction

Large Language Models (LLMs) excel in natural language understanding, yet their logical reasoning capabilities remain difficult to evaluate [1], [2], [3]. Logical fallacy detection has thus emerged as a key benchmark for assessing LLMs' ability to recognize flawed reasoning [4], [5], as it requires identifying faulty argumentative structures rather than merely parsing semantics.

Existing research on logical fallacy detection has largely focused on improving classification performance through prompt engineering, few-shot learning, and related inference-time strategies [6]. While these approaches often yield competitive results, their evaluation remains centered on prediction outcomes rather than on whether the model's explanation actually supports its label. As a result, correct predictions may derive from superficial correlations, with no guarantee that the accompanying explanation provides genuine support for the predicted label.

In parallel, recent studies have applied symbolic logic to represent argument structures more explicitly [3]. Nevertheless, these efforts formalize the input argument rather than the model's generated reasoning chain, and satisfiability checking does not establish that the conclusion follows from stated premises. What remains missing is an evaluation framework that jointly analyzes predicted labels, generated explanations, and formal verification outcomes.

Beyond methodological limitations, evaluation remains constrained by dataset construction and annotation. Annotation schemes compress interpretive variation into a limited set of labels, even when the same statement admits multiple readings [6], [7]. Label-based evaluation may therefore fail to capture reasoning that is consistent with an interpretation not reflected in the annotation.

Motivated by these challenges, we propose ForEx (Formal Verification for Explainable Reasoning), a framework that translates LLM-generated explanations into Lean4 and evaluates the formalized self-consistency of the translated rationale under encoded premises, rather than the validity or faithfulness of the original natural language argument. A successful Lean4 verification therefore certifies only that the translated conclusion is derivable from the encoded premises within the given formalization. By jointly examining prediction labels, translated rationales, and verification outcomes, ForEx provides a structured way to identify cases in which formally verified reasoning systematically diverges from existing annotations. Such cases may suggest that gold labels do not fully capture plausible interpretations of the input, making ForEx useful not only for reasoning verification but also for downstream annotation analysis and augmentation. In summary, the contributions of this paper are as follows:

1. We introduce ForEx, a framework that translates LLM-generated explanations into Lean4 and enables machine-checkable analysis of the resulting formalized reasoning chains.
2. We design the LLM Argument Verification Matrix, which separates label consistency from formal verification status and exposes mismatches between label correctness and formal derivability.

https://github.com/D3-Laboratory/ForEx

3. We develop a consensus-guided annotation pipeline that combines human annotation standards, multi-LLM agreement, and formal verification to support annotation analysis and conservative label augmentation.

The remainder of this paper is organized as follows: Section 2 reviews related work. Section 3 presents the ForEx framework, including the LLM Argument Verification Matrix and consensus-guided annotation pipeline. Section 4 reports experimental results on LOGIC-Climate. Section 5 concludes with future directions.

## II. Related Work

Prior work on logical fallacy detection has largely treated the task as supervised classification. Jin et al. [1] introduced the LOGIC dataset and LogicClimate challenge set with a structure-aware classifier. Yeh et al. [4] introduced CoCoLoFa and showed the effectiveness of supervised models for fallacy detection and classification. Lei and Huang [8] incorporated logical structure trees into LLMs to improve fallacy detection and classification. Despite these advances, prior work measures performance through label prediction, which is insufficient for assessing reasoning quality.

Complementing classification approaches, structured and symbolic methods encode fallacy definitions as relational graphs [9] or capture compositional argument structures through logical structure trees [8]. The most directly related is NL2FOL, which converts input arguments into first-order logic and checks satisfiability via SMT solvers [3]. Nevertheless, NL2FOL formalizes the input argument rather than the model's generated reasoning chain, and satisfiability checking does not show that the conclusion follows from the stated premises. These approaches therefore do not assess whether model-generated reasoning can be formally derived from encoded premises.

Another line of work targets reasoning reliability through behavioral verification. Chain-of-Verification (CoVe) prompts models to generate verification questions and revise their answers accordingly [10], while EVER performs step-wise hallucination detection and rectification during generation [11]. Generator–verifier studies further show that verification effectiveness depends strongly on verifier capability, generator strength, and problem difficulty [12]. However, these approaches evaluate linguistic coherence, factual consistency, or behavioral correctness, rather than whether a reasoning chain constitutes a formally valid derivation. As a result, a fluent but structurally fallacious rationale may still pass behavioral verification.

Compounding these limitations, annotation construction further highlights the limits of label-based evaluation. Standard annotation practice typically resolves disagreement through aggregation, such as majority voting or adjudication, treating disagreement primarily as noise. CrowdTruth 2.0 instead argues that disagreement can reflect ambiguity in the data and should be modeled as a signal rather than discarded [6]. Pavlick and Kwiatkowski [7] further show that disagreements in natural language inference often reflect multiple valid interpretations of the same text. This issue is especially relevant to fallacy detection, where judgments may depend on unstated assumptions and alternative reasoning paths. Compressing such cases into a single gold label can therefore obscure plausible interpretations and make evaluation overly dependent on label agreement rather than reasoning quality [13], [14]. Overall, existing approaches still do not directly evaluate whether a model's predicted label is supported by a reasoning process that can be explicitly represented and formally assessed. To address this gap, ForEx sits at the intersection of fallacy detection, argument formalization, and annotation-disagreement research: it translates model-generated explanations into Lean4 to assess formal derivability under encoded premises, while using systematic mismatches between verification outcomes and human labels to support annotation analysis and augmentation.

## III. Formal Verification for Explainable Reasoning

We present ForEx, a framework for assessing LLM-generated reasoning by linking natural language explanations to formal representations in Lean4. As illustrated in Figure 1, the workflow consists of three stages: Reasoning Candidates, Execution Feedback & Lean4 Verification, and Label Consistency Evaluation.

The framework begins with the Reasoning Candidates stage, where the Reasoner receives an input statement and generates up to $k$ candidates, where $k$ is a tunable upper-bound hyperparameter. Each candidate is represented as a tuple $c_i = (f_i, a_i, s_i)$, where $i \in \{1, \dots, k\}$, corresponding

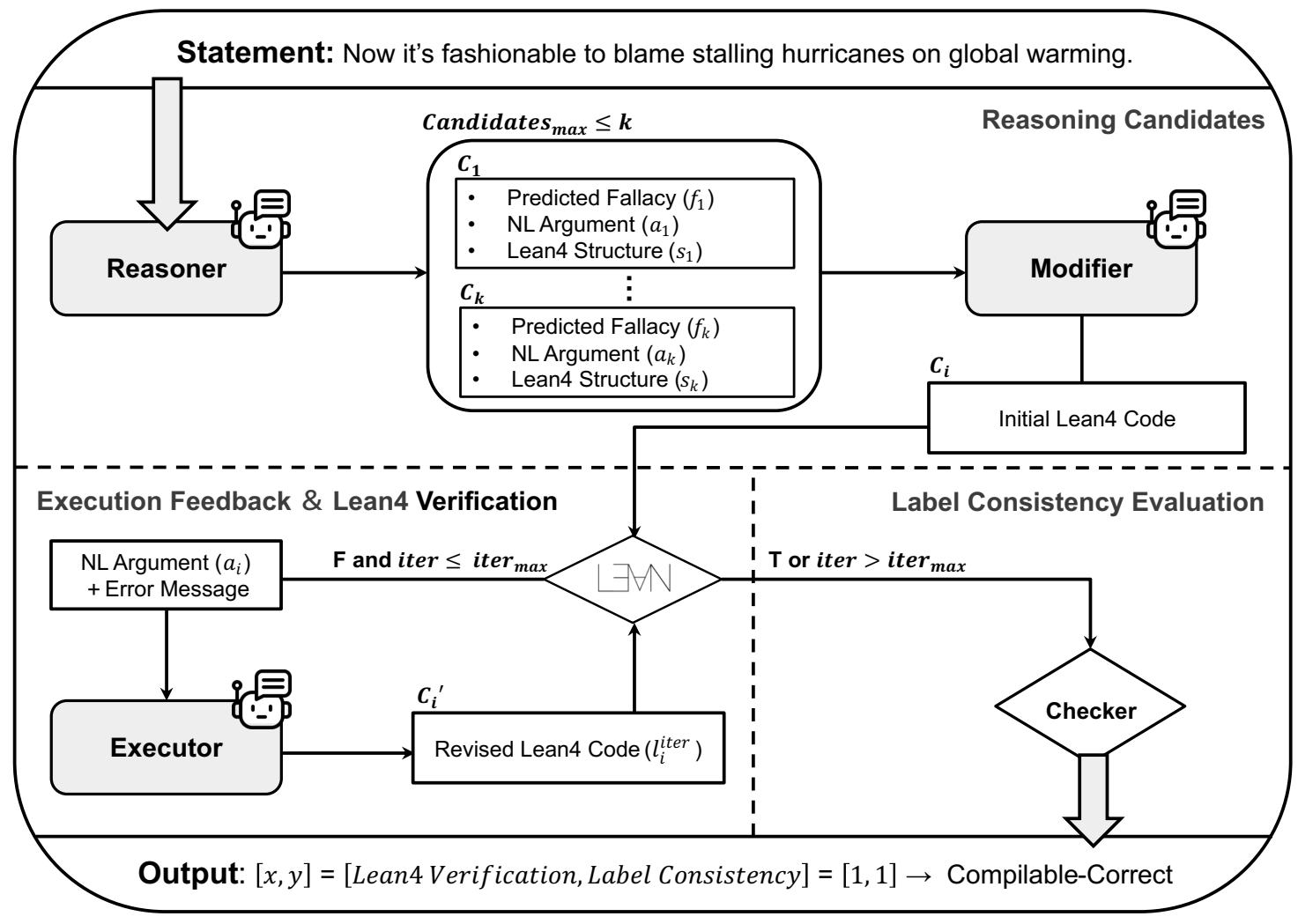

Fig. 1. Overview of the ForEx Framework

```
variable {Entity : Type}
variable (claim evidence : Entity)

variable (is_vague : Entity → Prop)
variable (lacks_specificity : Entity → Prop)
variable (draws_broad_conclusion : Entity → Prop)
variable (faulty_generalization : Entity → Prop)

axiom fg_rule :
  ∀ {c e : Entity}, is_vague e → lacks_specificity c →
draws_broad_conclusion c → faulty_generalization c

theorem example_fg
  (h1 : is_vague evidence)
  (h2 : lacks_specificity claim)
  (h3 : draws_broad_conclusion claim) :
  faulty_generalization claim :=
  fg_rule h1 h2 h3
```

Fig. 2. Example of Lean4 Verification

to the Predicted Fallacy, NL Argument, and Lean4 Structure shown in Figure 1. Here, $f_i$ denotes the predicted logical fallacy type, $a_i$ denotes the natural language argument explaining the reasoning process, and $s_i$ denotes the Lean4 structure used to formalize that explanation. These outputs are then passed to the Modifier, which converts each $s_i$ into Initial Lean4 Code for subsequent compilation and verification.

The second stage, Execution Feedback and Lean4 Verification, attempts to compile the generated Lean4 code. Let $iter$ denote the current repair iteration and $iter_{max}$ denote the maximum number of permitted repair steps. If the initial Lean4 code fails to compile (state F), the framework activates an execution-feedback loop while $iter \leq iter_{max}$. In each iteration, the Executor receives the compilation error message together with the original explanation $a_i$ and revises the Lean4 code accordingly. To reduce semantic drift, revisions are constrained to rely only on $a_i$ and the Initial Lean4 Code; the Executor is not allowed to introduce new logical content or alter the underlying rationale provided by the Reasoner. Each repair step produces an updated candidate $C_i'$ with Revised Lean4 Code ($l_i^{iter}$). The process continues until the code compiles successfully (state T) or the repair budget is exhausted. The resulting candidate is then forwarded to the Checker, which evaluates label consistency as described in the following section.

To clarify what “verification” means in our framework, we provide a simplified example of how a model-generated explanation is translated into Lean4. Consider the statement: *“There’s very strange things going on on planet Earth right now.”* A model may classify this statement as *Faulty Generalization* on the grounds that it draws a broad conclusion from vague and insufficiently specified evidence. This explanation can be formalized in Lean4 as a rule stating that if the cited evidence is vague, the claim lacks specificity, and the claim draws a broad conclusion, then the claim constitutes a faulty generalization, as shown in Figure 2. If Lean4 accepts the proof, this means only that the conclusion is derivable from the encoded premises and rule schema, not that the full meaning of the original natural language statement has been verified.

We then evaluate candidates in the Label Consistency Evaluation stage using the LLM Argument Verification Matrix shown in Table 1. The matrix separates label consistency from formal verification status, allowing us to analyze whether a model’s predicted label is accompanied by a reasoning chain that can be formalized and verified under the encoded premises. It is defined by two axes: Lean4 Verification and Label Consistency. Lean4 Verification indicates whether the revised Lean4 code compiles successfully in Lean4 (1 = Pass, 0 = Fail). Label Consistency indicates whether the predicted fallacy type matches the human annotation (1 = Match, 0 = Mismatch). Importantly, a verification pass means that the formalized reasoning chain is derivable under the encoded premises in Lean4; it does not by itself guarantee that the formalization fully captures the meaning or logical force of the original natural language statement. The two axes can be represented as a pair $[x, y]$, where $x$ denotes Lean4 Verification, and $y$ denotes Label Consistency. The four quadrants are defined as follows:

TABLE I. LLM ARGUMENT VERIFICATION MATRIX

| | | Label Consistency | |
|---|---|---|---|
| | | Match | Mismatch |
| Lean4 Verification | Pass | Compilable-Correct (Pass & Match) | Compilable-Alternative (Pass & Mismatch) |
| | Fail | Uncompilable-Correct (Fail & Match) | Uncompilable-Incorrect (Fail & Mismatch) |

- **Compilable-Correct Output [1, 1]**: The formalized reasoning chain compiles successfully in Lean4, and the prediction matches the annotation.
- **Compilable-Alternative Output [1, 0]**: The formalized reasoning chain compiles successfully in Lean4, but the prediction differs from the annotation, indicating a plausible alternative interpretation.
- **Uncompilable-Correct Output [0, 1]**: The formalized reasoning chain fails to compile in Lean4, but the prediction matches the annotation.
- **Uncompilable-Incorrect Output [0, 0]**: The formalized reasoning chain fails to compile in Lean4, and the prediction differs from the annotation.

Building on the LLM Argument Verification Matrix, we design a consensus-guided annotation pipeline to identify plausible additional labels under human-anchored, cross-model consensus. Because logical fallacy detection is treated as a multi-label task in our framework, each statement is represented as a set of fallacy labels rather than a single class assignment.

For latent behavioral analysis, each model is represented as a 14-dimensional feature vector over fallacy categories (13 fallacy types and one no-fallacy category). For each statement and model, a multi-hot vector is constructed in which each dimension indicates whether the corresponding label appears among the model’s verified outputs. If multiple verified labels are assigned, the corresponding dimensions are each set to 1 independently. Averaging these vectors across all evaluated statements yields a normalized behavioral signature for each model. Human annotations are encoded in the same space by averaging statement-level multi-hot label vectors across statements; unlike model predictions, they are not subject to Lean4 verification and thus capture label distribution rather than verified prediction behavior. To characterize global behavioral structure, we project these representations into two

TABLE II. CUMULATIVE PASS RATE (%) ACROSS REPAIR ITERATIONS FOR DIFFERENT REASONER MODELS

| Model (T/NT) | Modifier | 1st | 2nd | 3rd | 4th |
|---|---|---|---|---|---|
| Claude-sonnet-4.5 (T) | 14.63 | 94.31 | 100.0 | 100.0 | 100.0 |
| GPT-5 (T) | 47.89 | 93.07 | 98.79 | 99.99 | 99.99 |
| GPT-oss-120b (T) | 37.40 | 94.49 | 98.82 | 100.0 | 100.0 |
| Deepseek-chat-v3.1 (T) | 25.00 | 93.27 | 97.59 | 97.59 | 97.59 |
| Deepseek-r1 (T) | 39.91 | 92.96 | 98.59 | 98.59 | 98.59 |
| Kimi-k2 (T) | 49.51 | 94.75 | 97.70 | 98.69 | 99.02 |
| Gemini 2.5 flash (T) | 22.70 | 81.32 | 93.68 | 95.98 | 96.84 |
| Gemini 2.5 flash (NT) | 35.09 | 88.46 | 96.63 | 98.55 | 99.03 |
| Gemini 2.5 pro (T) | 22.70 | 81.32 | 93.68 | 95.98 | 96.84 |
| Gemini 2.5 pro (NT) | 24.48 | 85.31 | 95.10 | 96.50 | 96.85 |
| Grok 4 fast (T) | 38.40 | 90.72 | 97.47 | 98.73 | 98.73 |
| Grok 4 fast (NT) | 38.70 | 92.57 | 98.14 | 99.07 | 99.38 |
| Qwen3 235B (T) | 57.63 | 93.13 | 95.80 | 96.95 | 97.33 |
| Qwen3 235B (NT) | 69.86 | 95.21 | 97.26 | 97.94 | 97.94 |
| GPT-4o (NT) | 12.83 | 92.83 | 99.62 | 99.99 | 99.99 |
| **Average** | **36.70** | **91.11** | **97.37** | **98.39** | **98.65** |

**Note.** Values are percentages. T/NT denote thinking/non-thinking models; 1st–4th denote repair iterations.

dimensions via PCA, selected for simplicity and interpretability after observing consistent clustering across PCA, a linear autoencoder, and a non-negative autoencoder.

In the reduced space, low-consensus models are identified using Mahalanobis distance; those falling outside the 95% confidence ellipse are excluded from subsequent augmentation. After filtering, we quantify statement-level disagreement using pairwise Jaccard distance. For each statement $s$, let $\mathrm{A} = \{\mathrm{A}_1, \dots, \mathrm{A}_\mathrm{M}\}$ denote the label sets of all $M$ annotation sources (retained models and the human annotation). The disagreement score is defined as:

$$\mathrm{Score(s)} = \frac{1}{\binom{M}{2}} \sum_{i<j} \mathrm{JaccardDist}(\mathrm{A_i}, \mathrm{A_j}) \quad (1)$$

Averaging over all pairs ensures comparability even if the number of retained models varies. Statements at or below the first quartile are treated as high-consensus and retained for label augmentation.

For each retained statement, a label is added to the augmented annotation if at least 50% of retained models assign it under verification. Augmented annotations are then evaluated against the original human annotations using recall over statement-label pairs, defined as the proportion of original annotated labels recovered by the augmented annotation set.

## IV. EXPERIMENT AND DISCUSSION

We evaluate ForEx on the LOGIC-Climate dataset [1], which contains 1,074 instances across 13 fallacy categories. We sample 10% of the data (107 instances) as a representative subset for empirical evaluation, while preserving category balance. We tested 15 Reasoner models spanning both thinking and non-thinking variants. For non-thinking models, we apply Chain-of-Thought prompting [15] to externalize reasoning into a Lean4-verifiable format. To isolate Reasoner performance from Executor variability, we use Claude Sonnet 4.5 [16] as the Executor across all experiments, selecting it for its strong performance in structured code generation and reliable incorporation of compiler feedback. Across all experiments, we set $\mathrm{k} \leq 3$, corresponding to the maximum number of fallacy labels observed per instance, and set the maximum number of repair iterations to 4. Although the dataset is multi-label, most instances contain only one label (84.82% original, 71.96% sampled), which retains multi-fallacy cases while keeping the evaluation computationally manageable.

Table 2 reports pass rates across repair iterations. On average, 36.70% of Modifier outputs compile without repair, rising to 91.11% after one iteration and 98.65% by the fourth, indicating gains largely saturate after the first repair round. Within each model pair, thinking variants show lower pass rates than their non-thinking counterparts, likely because they generate more complex Lean4 structures that are harder to compile and repair. Remaining failures are categorized as Syntax Failure (SF).

Table 3 reports model performance under the LLM Argument Verification Matrix, which distinguishes formal verification status from label correctness. Quadrants are ordered by priority: Compilable-Correct (CC) > Compilable-Alternative (CA) > Uncompilable-Correct (UC) > Uncompilable-Incorrect (UI). CC+CA serves as the primary signal, capturing all predictions accompanied by a compilable reasoning chain regardless of label outcome. We evaluate predictions at the label level, assigning each predicted label to a matrix category to preserve information in the multi-label setting. Because models may output fewer

TABLE III. MODEL PERFORMANCE AND ERROR ANALYSIS UNDER THE ARGUMENT VERIFICATION MATRIX

| Model (T/NT) | Count | CC+CA↑ | CC↑ | CA↑ | UC↓ | UI↓ | VF↓ | NF | SF↓ |
|---|---|---|---|---|---|---|---|---|---|
| Claude-sonnet-4.5 (T) | 227 | 93.84 | **22.91** | 70.93 | 0.88 | 5.29 | 2.20 | 2.64 | 0.44 |
| GPT-5 (T) | 290 | **97.59** | 17.93 | **79.66** | 0.34 | **2.07** | 0.69 | 1.38 | **0.00** |
| GPT-oss-120b (T) | 234 | 97.01 | 20.51 | 76.50 | **0.00** | 2.99 | **0.00** | 2.99 | **0.00** |
| Deepseek-chat-v3.1 (T) | 193 | 90.16 | 20.21 | 69.95 | 1.04 | 9.33 | 1.04 | 5.70 | 2.59 |
| Deepseek-r1 (T) | 201 | 92.04 | 22.39 | 69.65 | 0.50 | 7.96 | **0.00** | 5.97 | 1.99 |
| Kimi-k2 (T) | 276 | 94.20 | 18.48 | 75.72 | 1.45 | 5.43 | 0.72 | 3.62 | 1.09 |
| Gemini 2.5 flash (T) | 229 | 89.52 | 21.40 | 68.12 | 2.62 | 7.86 | 6.11 | 0.44 | 1.31 |
| Gemini 2.5 flash (NT) | 195 | 90.77 | 22.05 | 68.72 | 1.03 | 8.21 | 3.08 | 4.10 | 1.03 |
| Gemini 2.5 pro (T) | 302 | 91.72 | 17.88 | 73.84 | 0.99 | 7.62 | 3.31 | 0.66 | 3.64 |
| Gemini 2.5 pro (NT) | 248 | 91.13 | 19.76 | 71.37 | 2.02 | 6.85 | 3.23 | 0.40 | 3.23 |
| Grok 4 fast (T) | 215 | 90.70 | 20.93 | 69.77 | 0.47 | 8.84 | 2.33 | 5.12 | 1.40 |
| Grok 4 fast (NT) | 279 | 96.41 | 17.92 | 78.49 | 1.08 | 2.87 | 1.79 | 0.36 | 0.72 |
| Qwen3 235B (T) | 237 | 92.40 | 20.25 | 72.15 | **0.00** | 7.59 | 2.11 | 2.95 | 2.53 |
| Qwen3 235B (NT) | 259 | 94.59 | 22.39 | 72.20 | 1.16 | 4.25 | 1.93 | 1.16 | 1.16 |
| GPT-4o (NT) | 236 | 96.61 | 22.88 | 73.73 | **0.00** | 3.39 | 0.85 | 2.54 | **0.00** |

**Note.** Values are percentages. T/NT denote thinking/non-thinking models; ↑ higher is better; ↓ lower is better. Best values are bold.

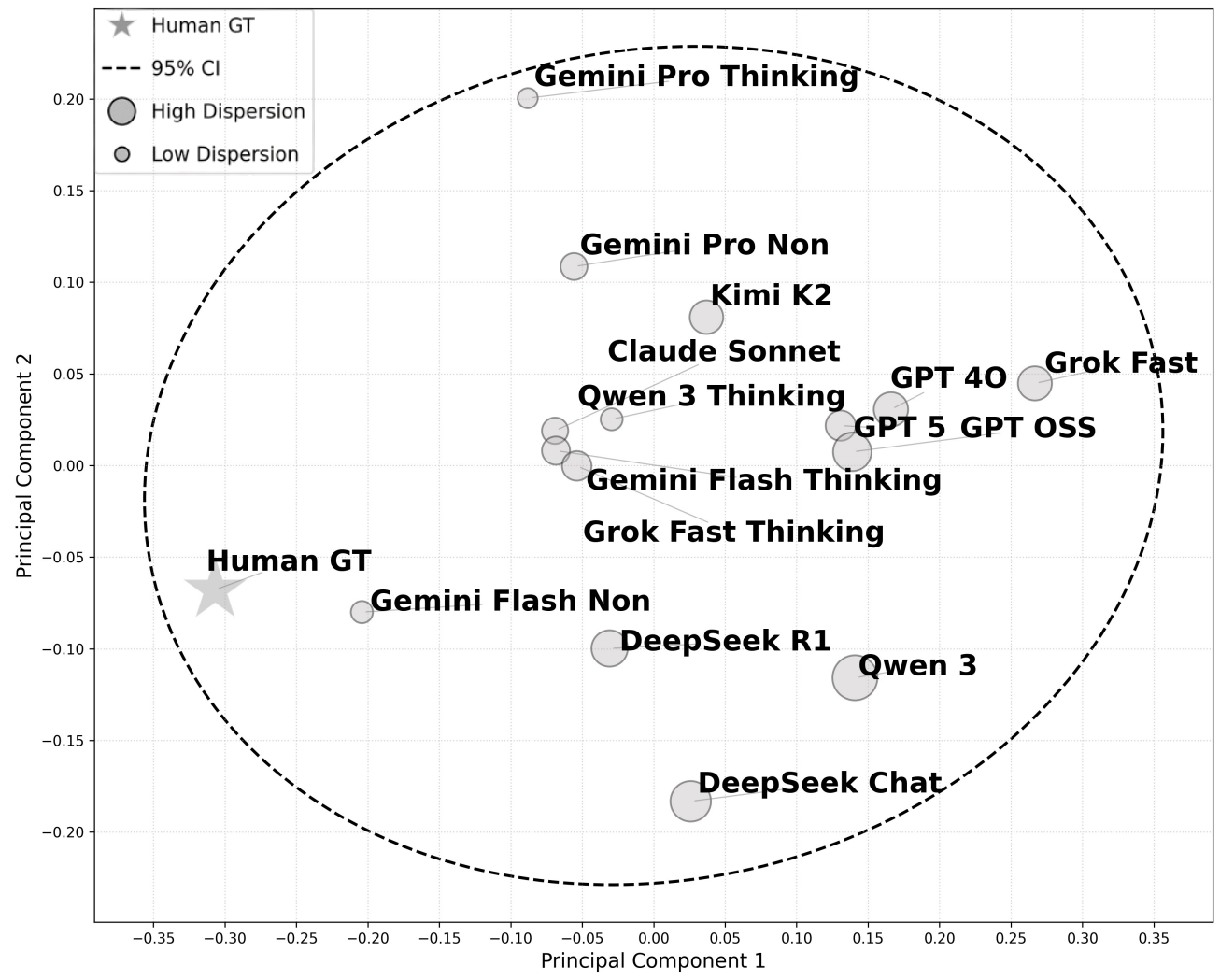

Fig. 3. PCA Model Clustering with Mahalanobis Distance

than $k$ labels for a given instance, Count denotes the number of evaluated label-level predictions for each model rather than the number of instances. The UI category is further decomposed into Verification Failure (VF), No Fallacy (NF), and Syntax Failure (SF) to distinguish between distinct failure modes.

The combined CC+CA metric is consistently high across models and typically exceeds 90%, with the top value reaching 97.59%. In contrast, CC scores show greater variation across models. Claude-sonnet-4.5 (22.91%), GPT-4o (22.88%) [17], and Deepseek-r1 (22.39%) [18] achieve the highest CC scores. This gap between CC and CC+CA indicates that many formally verified outputs fall into the CA category, pointing to plausible readings not captured by the annotated labels.

UC remains consistently low across models, generally below 2%, suggesting that correct labels rarely arise without a verifiable reasoning chain. Within UI, NF is the dominant failure mode, especially in the Deepseek series [18], [19]. One possible explanation is that isolated single-sentence inputs omit contextual information needed for reliable fallacy detection, which may also contribute to mismatches with human annotations. By contrast, VF and SF remain comparatively rare, with most values below 3% and 2%, respectively, indicating that failures are more often associated with label mismatch than with verification or syntax breakdown.

Motivated by the prevalence of CA outputs, we apply a consensus-guided annotation pipeline to identify additional supported labels. Figure 3 shows the PCA projection of the LLMs and the human annotation profile. Most models cluster near the human annotation profile, and no models are identified as outliers under the Mahalanobis distance criterion (95% confidence ellipse).

We next examine statement-level disagreement using the Jaccard-based score. The median distance across all statements is 0.63 (Q1: 0.56, Q3: 0.71). Statements at or below the first quartile threshold are treated as high-consensus and comprise approximately 25% of the dataset. Applying the 50% support criterion within this subset yields augmented annotations with 0.77 recall against the original human annotations, indicating that most original labels are preserved by the augmented set. The newly added labels are formally supported and retained through consensus-guided filtering, but their status as gold annotations remains to be confirmed.

The following discussion distinguishes observed patterns from their possible interpretations. First, the prevalence of CA outputs suggests that many formally supported predictions may reflect plausible alternative interpretations rather than simple errors (Appendix, Case 1), though this interpretation remains tentative. Second, NF predictions often appear in context-limited cases. Prior analysis of LOGIC-Climate shows that 40% of errors stem from isolated arguments [20], and Appendix Case 2 illustrates how some NF predictions can appear plausible under limited-context reading, possibly because annotators had access to broader discourse context not available to the models. Third, the limited number of consensus augmentations (27 statements) reflects the strictness of our criterion, which retains only interpretations supported by multiple LLMs and the human annotation. Fourth, the low UC rate (<2%) indicates that correct labels rarely co-occur with unverified reasoning chains in our setting.

While these findings provide insights into LLM reasoning, several limitations warrant discussion. The plausibility of CA predictions cannot be confirmed without clearer evidence about annotator methodology. In addition, formal verification does not guarantee semantic alignment between the formalized reasoning and the original natural language statement. Under limited context, derivable reasoning chains may still be misaligned with human interpretation. More broadly, because natural language lacks a fully specified semantic foundation, formal verification results may overstate model capability. Finally, NF errors within UI may arise from either insufficient semantic understanding or loss of contextual information, and the current framework does not distinguish between these explanations. Regarding the annotation pipeline, cases exhibiting systematic model-human divergence may be filtered out by the augmentation criterion (Appendix, Case 3). The augmentation results should therefore be interpreted as a demonstration of conservative, consensus-guided label expansion rather than evidence of annotation quality improvement.

Additional case-level examples, including model predictions, Lean4 translations, verification outcomes, and annotation-consensus results, are available in the project repository: https://github.com/D3-Laboratory/ForEx

## V. Conclusion

This paper addresses a key challenge in logical fallacy detection: evaluation is typically centered on label correctness, while the formal status of the supporting reasoning is rarely assessed separately. To address this limitation, we propose ForEx, a framework that links natural language explanations to Lean4 representations, along with an LLM Argument Verification Matrix that decouples label correctness from formal verification status. We further introduce a consensus-guided annotation pipeline for analyzing agreement patterns between human annotations and multi-LLM predictions.

Experiments demonstrate that formal verification can be used to assess whether model-generated reasoning chains are derivable under encoded premises. The results also reveal a substantial gap between verification outcomes and agreement with existing human labels. These findings suggest that label-based evaluation and formal verification capture different aspects of model behavior. Future work will focus on improving the alignment between formal verification outcomes and semantic correctness by extending the framework to richer argument structures, incorporating contextual information, and integrating semantic consistency checks.

## Acknowledgement

This work was supported by the Taiwan Comprehensive University System (TCUS).

## Appendix. Case Analysis Examples

### Case 1. Label Mismatch Under Limited Context

- **Statement:** For good reason: In the U.S., and for much of the world, the most dangerous environmental pollutants have been cleaned up. U.S. emissions of particulates, metals, and varied gases, all of these: ozone, lead, carbon monoxide, oxides of nitrogen, and sulfur, fell almost 70 % between 1970 and 2014.
- **Model Prediction:** Faulty Generalization (15/15 models)
- **Human Annotation:** Ad Hominem
- **Analysis:** All models predicted Faulty Generalization based on the isolated statement. The human annotation reflects a different interpretation from the model predictions. One possible explanation is that the isolated sentence omits contextual information relevant to fallacy identification, suggesting that single-sentence inputs may provide insufficient cues for reliable classification.

### Case 2. No Fallacy (NF)

- **Statement:** Barrie now works for the Climate Change Institute at Australian National University, Canberra.
- **Model Prediction:** No Fallacy (14/15 models)
- **Human Annotation:** Intentional Fallacy
- **Analysis:** The statement appears factually neutral in isolation, leading most models to predict No Fallacy. The mismatch with the human annotation suggests that the assigned label may depend on information not recoverable from the single sentence alone, illustrating the difficulty of fallacy detection under limited context.

### Case 3. Systematic Label Mismatch Under Argument-Level Reasoning

- **Statement:** There are no carbon emissions, because carbon is a solid and most carbon is black and soot, which are solids.
- **Model Prediction:** Equivocation (12/15 models), False Causality (11/15 models)
- **Human Annotation:** Fallacy of Logic
- **Analysis:** Most models predicted Equivocation or False Causality based on local cues such as the repeated use of "carbon" or the apparent causal structure. In contrast, the human label, Fallacy of Logic, reflects a broader argument-level judgment that the inference is invalid. This suggests that models may rely on local cues rather than assessing overall argument validity. The case was excluded from label augmentation due to high Jaccard disagreement, showing that the pipeline can filter out high-divergence model–human mismatches.